\documentclass[letterpaper]{article} 
\usepackage[]{aaai2026}  
\usepackage{times}  
\usepackage{helvet}  
\usepackage{courier}  
\usepackage[hyphens]{url}  
\usepackage{graphicx} 
\usepackage{natbib}  
\usepackage{caption} 
\usepackage{algorithm}
\usepackage{algorithmic}
\usepackage{multirow}
\usepackage{soul} 
\setul{0.15ex}{0.3ex}
\usepackage{makecell}
\usepackage{amssymb}
\usepackage{amsmath}
\usepackage{amsmath}

\usepackage{newfloat}
\usepackage{listings}
\DeclareCaptionStyle{ruled}{labelfont=normalfont,labelsep=colon,strut=off} 
\floatstyle{ruled}
\newfloat{listing}{tb}{lst}{}
\floatname{listing}{Listing}
\nocopyright

\title{The Geometry of Dialogue: Graphing Language Models to Reveal Synergistic Teams for Multi-Agent Collaboration}
\author{
    Kotaro Furuya\textsuperscript{\rm 1}, Yuichi Kitagawa\textsuperscript{\rm 1}
} 
\affiliations{
    \textsuperscript{\rm 1} Research and Development Group, Hitachi, Ltd.\\

    1-280 Higashi-Koigakubo, Kokubunji-shi,\\
    Tokyo 185-8601, Japan\\
    kotaro.furuya.ao@hitachi.com
}

\makeatletter
\def\blfootnote{\gdef\@thefnmark{}\@footnotetext}
\makeatother

\begin{document}

\maketitle
\blfootnote{Accepted at the AAAI-26 Workshop on LaMAS 2026 (Oral).}

\begin{abstract}
While a multi-agent approach based on large language models (LLMs) represents a promising strategy to surpass the capabilities of single models, its success is critically dependent on synergistic team composition. However, forming optimal teams is a significant challenge, as the inherent opacity of most models obscures the internal characteristics necessary for effective collaboration. In this paper, we propose an interaction-centric framework for automatic team composition that does not require any prior knowledge including their internal architectures, training data, or task performances. Our method constructs a "language model graph" that maps relationships between models from the semantic coherence of pairwise conversations, and then applies community detection to identify synergistic model clusters. Our experiments with diverse LLMs demonstrate that the proposed method discovers functionally coherent groups that reflect their latent specializations. Priming conversations with specific topics identified synergistic teams which outperform random baselines on downstream benchmarks and achieve comparable accuracy to that of manually-curated teams based on known model specializations. Our findings provide a new basis for the automated design of collaborative multi-agent LLM teams.
\end{abstract}

\begin{figure*}[t]
\centering
\includegraphics[width=0.9\textwidth]{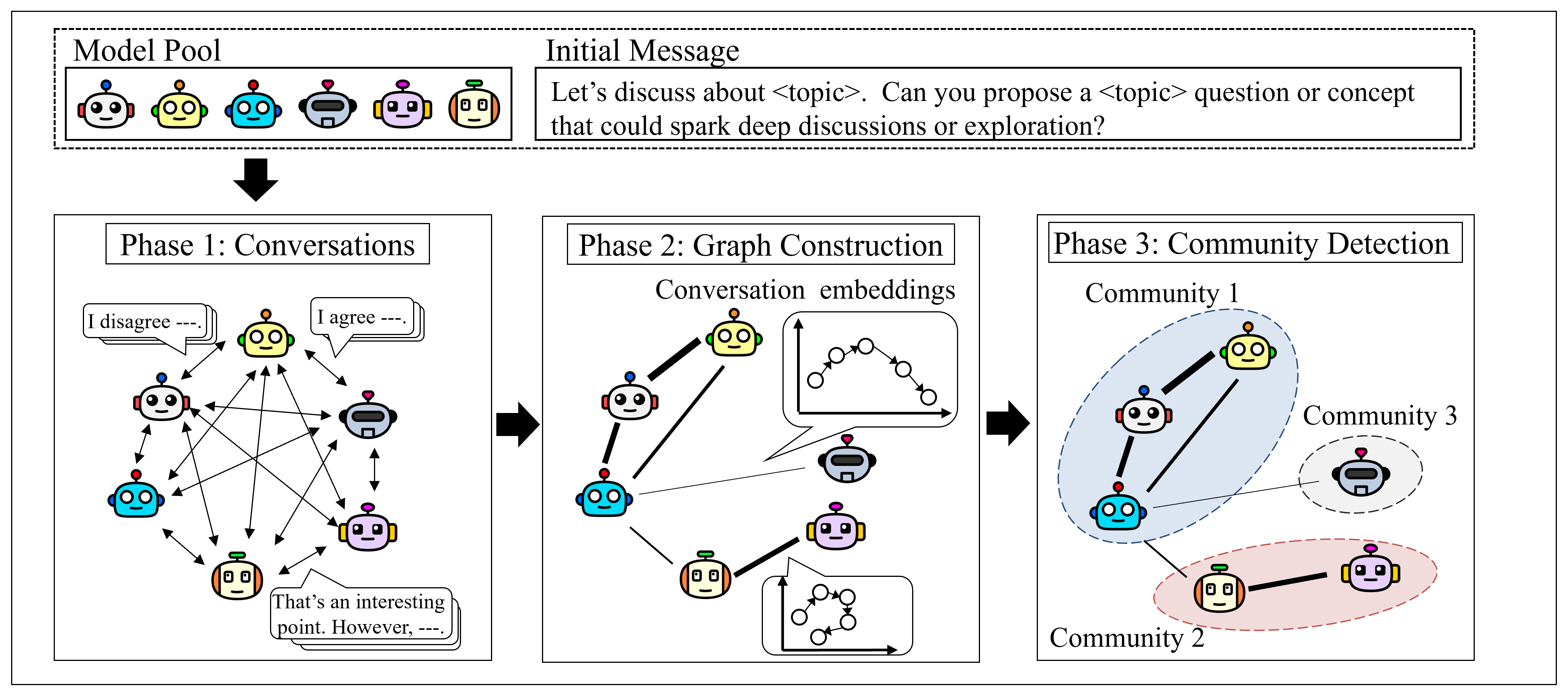} 
\caption{Overview of the proposed method. The method consists of three main phases: (1) generating conversations between pairs of language models, (2) constructing a language model graph based on the embeddings of their conversations, and (3) extracting model clusters via community detection.}
\label{fig:overview}
\end{figure*}

\section{Introduction}
Large language models (LLMs) have demonstrated remarkable capabilities across a wide range of natural language tasks, driven by advances in model scaling \cite{zhao2023survey}. 
However, the monolithic approach of scaling a single, massive model faces challenges such as hallucination \cite{maynez2020faithfulness} and catastrophic forgetting \cite{kirkpatrick2017overcoming}, limiting its ability to achieve robust performance across all domains.
In response, a multi-agent approach based on LLMs has emerged as a powerful paradigm \cite{tran2025multi}.
By coordinating multiple specialized or diverse models, this approach can mitigate the weaknesses of individual agents, leading to superior performance and robustness.
For instance, a team of agents with distinct roles can debate to refine reasoning, reflect on solutions to mitigate errors, or pool diverse knowledge to enhance reliability \cite{du2023improving, chen2023reconcile}.

However, the success of such cooperative strategies is critically dependent on the composition of the model team \cite{gu2025agentgroupchat}.
An effective team requires a synergistic combination of agents matched to the task.
Naively assembling high-performing models can stifle cognitive diversity \cite{li2025assessing}, while including underperforming models can degrade the collective output. 
Thus, the central challenge in effective multi-agent collaboration is identifying the optimal team structure by considering the task, the model characteristics, and the latent relationships between them to realize a synergistic combination.

Current research into automated team composition has predominantly adopted a task-driven, top-down approach to address this challenge \cite{chen2023autoagents, liu2024dynamic, song2025adaptive}. 
These frameworks begin by analyzing the functional requirements of a defined task and then proceed to recruit or assign agents to fulfill specific roles.
While powerful, this task-centric approach operates on a shared principle: the task dictates the team's design.
Consequently, this perspective naturally prioritizes task-specific agent selection, placing less emphasis on the underlying, latent relationships between models. 
Exploring these intrinsic synergies is essential for forming more versatile and robust collaborative teams.

However, a significant barrier to such an understanding is the opacity of LLMs. 
Ideally, team selection would be informed by knowledge of each model’s architecture, size, and training data. 
In practice, many leading commercial models are accessible only via APIs, which conceal their internals.
Even for open models, often only the model weights are released, while the complete training data remains undisclosed. 
Evaluating performance on downstream benchmarks is an alternative to explore model characteristics, but this is often impractical, requiring test data for every potential task and limiting applicability to novel domains.
This lack of transparency makes it difficult to assess a model's intrinsic strengths and potential synergies with others in advance.

In this paper, we propose an interaction-centric methodology for automated team composition, departing from the task-centric paradigm. 
To map the latent relational structure among LLMs, our method generates a corpus of pairwise conversations and represents their interactions within a "language model graph." 
In this network, the semantic coherence between models forms edge weights. 
Subsequent analysis using community detection algorithms then identifies clusters of models with high mutual affinity, revealing promising collaborative candidates.

Our primary contributions are:
\begin{enumerate}
    \item We propose an automatic LLM team composition framework based on inter-model conversations, which does not require any prior knowledge of their internal architecture, training data, or external test data.
    \item We empirically demonstrate that our graph-based method can identify clusters of models consistent with their latent characteristics.
    \item We validate our approach through performance evaluation on downstream benchmarks, showing that teams constructed from our identified clusters achieve higher accuracy than randomly assembled teams and approach the performance of manually-curated teams grouped by their known specializations.
\end{enumerate}

\section{Methodology}

\begin{figure*}[t]
\centering
\includegraphics[width=0.9\textwidth]{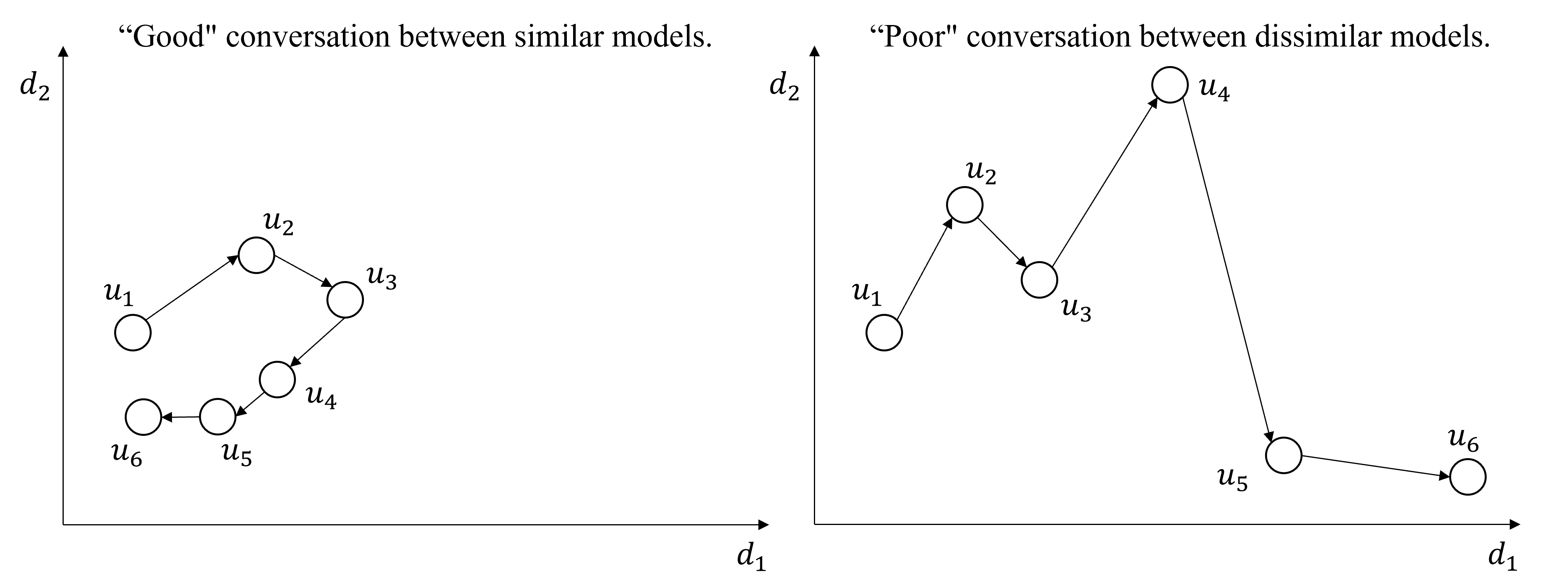} 
\caption{Illustration of the core idea. The left panel shows a "good" conversation between two similar models, where their utterances remain semantically close in the embedding space. The right panel depicts a "poor" conversation between two dissimilar models, where their utterances diverge significantly.}
\label{fig:embedding}
\end{figure*}

\subsection{Overview}
Our approach is to construct a language model graph that represents the relationships between models and then apply community detection algorithms to extract clusters.

This methodology is inspired by social graphs, where individuals and relationships are modeled as nodes and edges to identify communities \cite{falahi2010social}.
In our paradigm, language models serve as the nodes, and their interactions are quantified as weighted edges. 
We hypothesize that models with similar knowledge and capabilities will form densely connected clusters, while poorly performing or dissimilar models will appear as isolated nodes. 
This graph-based representation allows for the selection of promising model ensembles for collaboration while simultaneously filtering out those likely to degrade collective performance.

As illustrated in Figure \ref{fig:overview}, our method comprises three phases. 
In Phase 1 (Conversation Generation), we systematically generate pairwise conversations among a set of target models. 
In Phase 2 (Graph Construction), we analyze the semantic content of these dialogues to compute relationship scores, which then serve as edge weights to build the language model graph. 
Finally, in Phase 3 (Community Detection), we apply a community detection algorithm to the graph and identify clusters of models, which represent promising teams for collaborative tasks.

\subsection{Core Idea and Assumptions}
Before introducing the specifics of our method, we outline the foundational core idea of our approach.
Our methodology is predicated on two fundamental assumptions:
\begin{enumerate}
    \item Constructive dialogues unfold within a coherent semantic space.
    \item Language models with similar characteristics are more likely to engage in constructive dialogues.
\end{enumerate}

The first assumption is rooted in principles of effective human communication, specifically the linguistic concepts of common ground \cite{brown2016memory}.
Common ground refers to the information that is mutually known between the discourse partners, which provides the basis for an efficient exchange of ideas.
Productive discussions build upon this shared context by maintaining a logical and topical focus, which distinguishes a meaningful conversation from a random sequence of sentences. 
In contrast, conversations with misaligned topics or disparate background knowledge lack this shared foundation, often leading to misunderstandings and a failure to reach a constructive outcome.

The second assumption posits that models with similar features, such as training data or architecture, can better interpret each other's outputs, leading to collaborative dialogue.
This similarity of models provides an analogy to the common ground necessary for effective communication. 
Thus, similar models can share a common ground, leading to more consistent responses and effective communication.

Building on these assumptions, we hypothesize that the quality of a conversation between two models is reflected in the geometric properties of their utterances in embedding spaces. 
As depicted on the left in Figure \ref{fig:embedding}, a "good" conversation between similar models is expected to trace a dense trajectory, with their outputs remaining in close proximity. 
In contrast on the right, a "poor" conversation between dissimilar models will trace a sparse and distant trajectory as their outputs diverge. 
We therefore posit that the geometric relationships between embedded utterances can serve as a proxy for both conversational quality and the functional similarity between models. 
This allows us to quantify inter-model relationships using metrics like cosine similarity, revealing optimal collaborative groupings from their outputs alone, without requiring access to their internal states.

\subsection{Phase 1: Conversation Generation}
In this phase, we generate conversational data for all unique pairs of $N$ language models within a given set $\mathcal{M} = \{M_1, \dots, M_N\}$. 
The set of all unique pairs is denoted by $\mathcal{S} = \{ (M_i, M_j) \mid 1 \le i < j \le N \}$.

For each pair $(M_i, M_j) \in \mathcal{S}$, a single conversation is generated. 
All models operate under a common system prompt, $P_{\mathrm{sys}}$. 
One model is randomly selected to be the designated first speaker. 
The conversation begins by setting the first utterance, $u_1$, to a predefined starting message, $P_{\mathrm{start}}$. 
Thus, $u_1 = P_{\mathrm{start}}$. 
This initial utterance is then passed as input to the second model, which generates the second utterance, $u_2$. 
From this point, the two models continue to exchange utterances in a turn-based manner. 
The dialogue continues for a maximum of $K_{\mathrm{max}}$ turns or until either model generates a special termination token, $T_{\mathrm{stop}}$.
A turn is defined as a sequence of two utterances, one from each model.

The resulting conversation for a pair $(M_i, M_j)$ is recorded as a chronologically ordered sequence of utterances, $C_{i,j} = (u_1, u_2, \dots, u_L)$, where $L$ is the total number of utterances exchanged. 
For the subsequent analysis, we only consider complete turns, which are pairs of consecutive utterances $(u_{2k-1}, u_{2k})$ for $k \ge 1$.

\subsection{Phase 2: Language Model Graph Construction}

\subsubsection{Relationship Value Calculation}
From each conversation history $C_{i,j}$, we compute a relationship value, $r_{i,j}$, that quantifies the semantic similarity of their interaction.

First, we use a pre-trained sentence embedding function, $f: \mathcal{U} \to \mathbb{R}^d$, to map each utterance $u \in \mathcal{U}$, where $\mathcal{U}$ is the space of all possible utterances, to a $d$-dimensional vector.

The relationship $r_{i,j}$ is defined as the sum of cosine similarities between the vector embeddings of the utterance pairs from each complete turn. 
Let $K_{i,j} = \lfloor |C_{i,j}| / 2 \rfloor$ be the number of complete turns in the conversation $C_{i,j}$. 
The relationship is then calculated as:
\begin{equation*}
r_{i,j} = \sum_{k=1}^{K_{i,j}} \mathrm{sim}_{\cos}(f(u_{2k-1}), f(u_{2k})), 
\end{equation*}
where $u_{2k-1}$ and $u_{2k}$ are the utterances in the $k$-th complete turn, and $\mathrm{sim}_{\cos}(\mathbf{a}, \mathbf{b}) $ is the cosine similarity between two vectors $\mathbf{a}$ and $\mathbf{b}$. 
This process is repeated for all pairs $(M_i, M_j) \in \mathcal{S}$ to obtain the set of all relationships.

\subsubsection{Graph Construction}
We construct an undirected, weighted graph $G = (\mathcal{M}, \mathcal{E}, r)$ to represent the significant relationships among the language models. 
The set of vertices $\mathcal{M}$ consists of the language models. 
To focus our analysis on the most substantial connections, the edge set $\mathcal{E}$ includes only those pairs of models $(M_i, M_j)$ whose relationship value $r_{i,j}$ meets or exceeds a predefined threshold $\tau$. 
The weight of each edge is its corresponding relationship value.
Thus, the edge set $\mathcal{E}$ is formally defined as:
\begin{equation*}
\mathcal{E} = \{ (M_i, M_j) \in \mathcal{S} \mid r_{i,j} \ge \tau \}.
\end{equation*}
This method filters out weak or noisy interactions, thereby constructing a sparse graph that highlights the strongest relational structures. 
The threshold $\tau$ is a tunable hyperparameter that controls the sparsity of the graph.

\subsection{Phase 3: Cluster Extraction via Community Detection}
In the final phase, we analyze the language model graph $G$ using a community detection algorithm to identify densely connected subgroups of models. 
These communities represent clusters of models that are functionally similar and likely to collaborate effectively.

For this task, we employ the Louvain method, a widely-used hierarchical clustering algorithm known for its efficiency and effectiveness \cite{blondel2008fast}. 

\section{Experiments}
We conducted a series of experiments to validate our proposed method. 
First, we tested whether the language model graph could reveal latent model specializations by evaluating its ability to cluster models according to their known capabilities. 
Second, to validate the effectiveness of our team composition approach, we benchmarked the collective downstream task performance of teams constructed from the identified clusters against several baselines.

\subsection{Models and Implementation}
\begin{table}[]
\begin{tabular}{l|l|l}
\hline
\# & Model                                & Type  \\ \hline
1  & mistralai/Mistral-7B-Instruct-v0.2   & (i)   \\
2  & meta-llama/Meta-Llama-3-8B-Instruct  & (i)   \\
3  & Qwen/Qwen2-7B-Instruct               & (i)   \\
4  & Qwen/Qwen2-0.5B-Instruct             & (ii)  \\
5  & google/gemma-3-1b-it                 & (ii)  \\
6  & mistralai/Mathstral-7B-v0.1          & (iii) \\
7  & Qwen/Qwen2-Math-7B-Instruct          & (iii) \\
8  & BioMistral/BioMistral-7B             & (iv)  \\
9  & ContactDoctor/Bio-Medical-Llama-3-8B & (iv)  \\
10 & google/medgemma-4b-it                & (iv)  \\ \hline
\end{tabular}
\caption{List of models.}
\label{tab:models}
\end{table}

\begin{table}[]
\begin{tabular}{l|p{6.2cm}}
\hline
Parameter & Value                                                                                                                                                                                                                                                                                                                                                                                                                 \\ \hline
$P_\mathrm{start}$  & (a) Let's discuss a topic of your expertise or interest. Can you propose a question or concept that could spark deep discussions or exploration? \newline (b, c) Let's discuss about \textless topic\textgreater.  Can you propose a \textless topic\textgreater~question or concept that could spark deep discussions or exploration?\\
$P_\mathrm{sys}$ & Continue the debate, each response should be concise and as negative or critical as possible, while remaining logically sound. Show evidence of your reasoning and avoid repeating the same point unless absolutely necessary. If you feel the discussion has exhausted the available information and if the conversation becomes unclear, inappropriate, or non-contributive responses, please say `END DISCUSSION'. \\
$T_\mathrm{stop}$  & END DISCUSSION                                                                                                                                                                                                                                                                                                                                                                                                        \\
$K_\mathrm{max}$   & 5                                                                                                                                                                                                                                                                                                                                                                                                                     \\ 

$\tau$             & Median of all relationship values                                                                                                                                                                                                                                                                                                                                                                                            \\
\hline
\end{tabular}
\caption{Parameters.}
\label{tab:params}
\end{table}

Our experiments were conducted using a diverse set of ten language models, detailed in Table \ref{tab:models}. 
The collection includes four types: (i) relatively small-scale, general-purpose models \cite{jiang2023mistral7b,grattafiori2024llama,yang2024qwen2}; (ii) small-scale, general-purpose models \cite{yang2024qwen2,gemma_2025}; (iii) models fine-tuned for mathematics \cite{yang2024qwen2,jiang2023mistral7b}; and (iv) models fine-tuned for the medical domain \cite{labrak2024biomistral,ContactDoctor_Bio-Medical-Llama-3-8B,sellergren2025medgemma}. 
All models were served using the vLLM \cite{kwon2023efficient} and the multi-agent conversations were managed by the AutoGen package \cite{wu2023autogen}.

Key hyperparameters are listed in Table \ref{tab:params}. 
To investigate the effect of topical context, we initiated conversations using three distinct prompts: (a) a general, open-ended prompt; (b) a mathematics-focused prompt; and (c) a medical-focused prompt.
The system prompt was designed to elicit critical debate and mitigate model sycophancy \cite{sharma2024towards}.
Following generation, conversations were embedded using the multilingual-e5-large \cite{wang2024multilingual}. 
We set threshold $\tau$ to the median of all relationship values to focus on the significant interactions.
To minimize sequential bias, relationship values were calculated by averaging the outcomes of five conversation runs for each model pair.
Finally, communities were identified using the Louvain algorithm, as implemented in the NetworkX package \cite{networkx}. 

\begin{table*}[]
\centering 
\begin{tabular}{l|l|p{12cm}} 
\hline
$P_{\mathrm{start}}$ & Com. \# & Models \\ 
\hline
(a) General & 1 & Mistral-7B-Instruct-v0.2 (i), Meta-Llama-3-8B-Instruct (i), Qwen2-7B-Instruct (i), BioMistral-7B (iv), medgemma-4b-it (iv) \\
 & 2 &  Qwen2-0.5B-Instruct (ii), Mathstral-7B-v0.1 (iii), Qwen2-Math-7B-Instruct (iii), Bio-Medical-Llama-3-8B (iv)\\
 & 3 & gemma-3-1b-it (ii)\\ 
\hline
(b) Mathematical & 1 & Mistral-7B-Instruct-v0.2 (i) , Meta-Llama-3-8B-Instruct (i), BioMistral-7B (iv), Bio-Medical-Llama-3-8B (iv) \\
 & 2 & Qwen2-7B-Instruct (i), Qwen2-0.5B-Instruct (ii), Mathstral-7B-v0.1 (iii), Qwen2-Math-7B-Instruct (iii)\\
 & 3 & gemma-3-1b-it (ii), medgemma-4b-it (iv) \\
\hline
(c) Medical & 1 & Meta-Llama-3-8B-Instruct (i), Qwen2-7B-Instruct (i), Qwen2-0.5B-Instruct (ii), Mathstral-7B-v0.1 (iii), BioMistral-7B (iv)\\
 & 2 & Mistral-7B-Instruct-v0.2 (i), medgemma-4b-it (iv), Bio-Medical-Llama-3-8B (iv) \\
 & 3 & gemma-3-1b-it (ii) \\ 
\hline
\end{tabular}
\caption{Detected model communities for each topic. The parentheses indicate the model type listed in Table \ref{tab:models}.}
\label{tab:community_results}
\end{table*}

\subsection{Evaluation Setup}
To evaluate the performance of our method, we used a suite of benchmark datasets under a zero-shot inference setting.
We selected subsets of MMLU \cite{hendrycks2021measuring} to assess mathematical ability (abstract\_algebra, college\_mathematics, college\_physics) and medical knowledge (clinical\_knowledge, college\_biology, college\_medicine).
We further assessed mathematical and medical expertise using GSM8K \cite{cobbe2021gsm8k} and MATH-500 \cite{lightman2023lets}, and MedQA \cite{jin2021disease} and MedMCQA \cite{pmlr-v174-pal22a} respectively.
For latter reasoning-intensive datasets, we utilized a Chain-of-Thought (CoT) prompt \cite{wei2023chainofthoughtpromptingelicitsreasoning}, with an exception for BioMistral-7B and Bio-Medical-Llama-3-8B.
CoT prompting was omitted for these two models as it led to a collapse in their responses.
Performance was measured by overall accuracy on each dataset.
The collective answer for a team was determined by a majority vote on the final answers from each model.
To ensure robustness, all reported scores are the average of five independent trials.
We benchmarked the performance of our community-based teams against four baselines:\\
\textbf{Single-model}: The performance of the best-performing model, Qwen2-7B-Instruct with self-consistency \cite{wang2023selfconsistency}. For each question, the model generated ten independent responses, and the most frequent answer was selected as the final output.\\
\textbf{All-models}: The collective performance of all ten models voting together. This baseline measures the performance of a naive, non-selective ensemble.\\
\textbf{Random@3models}: The average performance of teams composed of three randomly selected models. The reported score is the average accuracy over five randomly composed teams. This represents the expected performance without any intelligent team selection.\\
\textbf{Type-based}: Teams constructed by grouping models according to their predefined specialization type as listed in Table \ref{tab:models}. This baseline serves as a practical upper bound, representing the performance achievable with prior knowledge.\\

\begin{figure}
\centering
\includegraphics[width=0.48\textwidth]{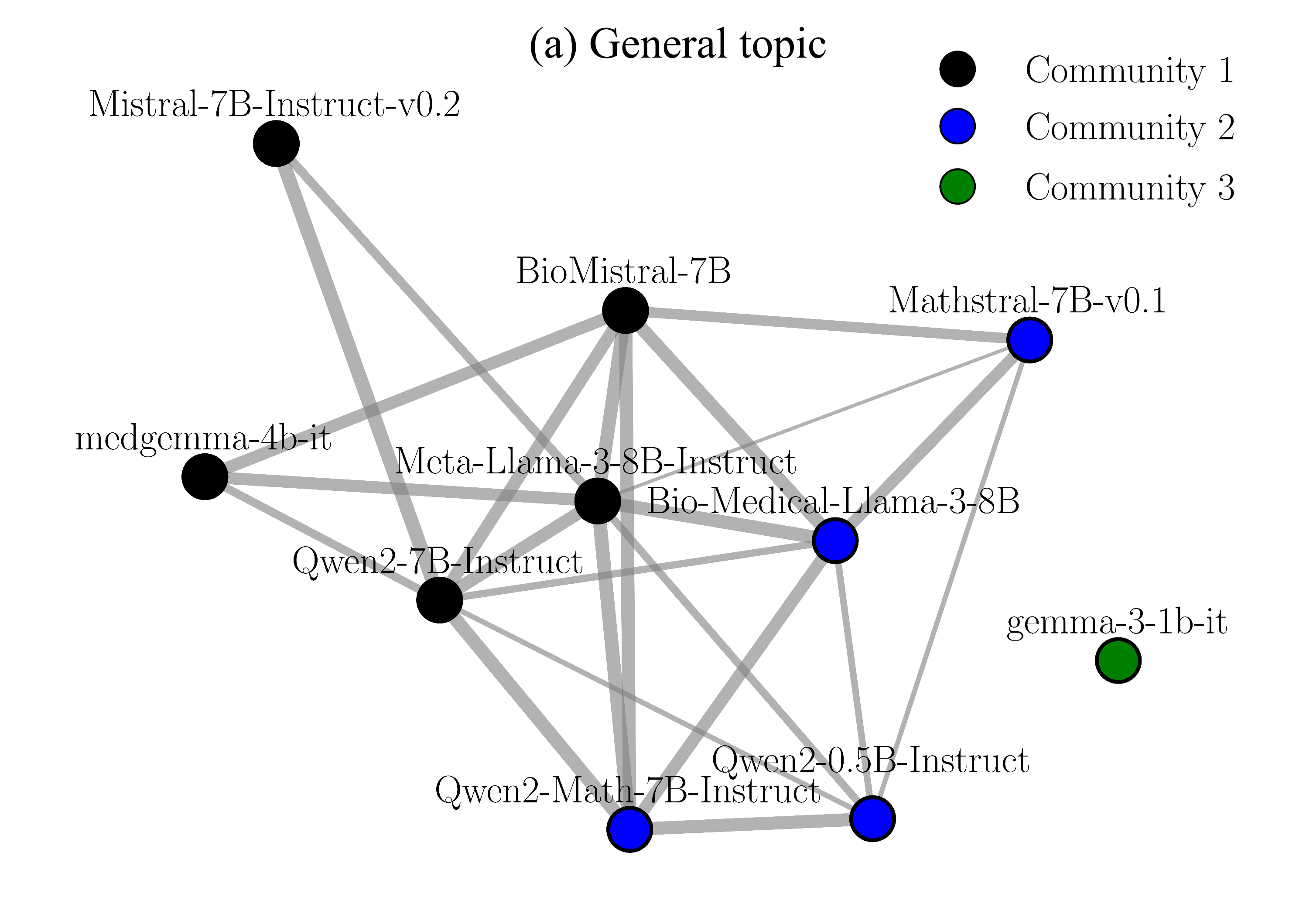} 
\caption{Language model graph constructed under the (a) general topic condition.}
\label{fig:graph_a}
\end{figure}

\begin{figure}
\centering
\includegraphics[width=0.48\textwidth]{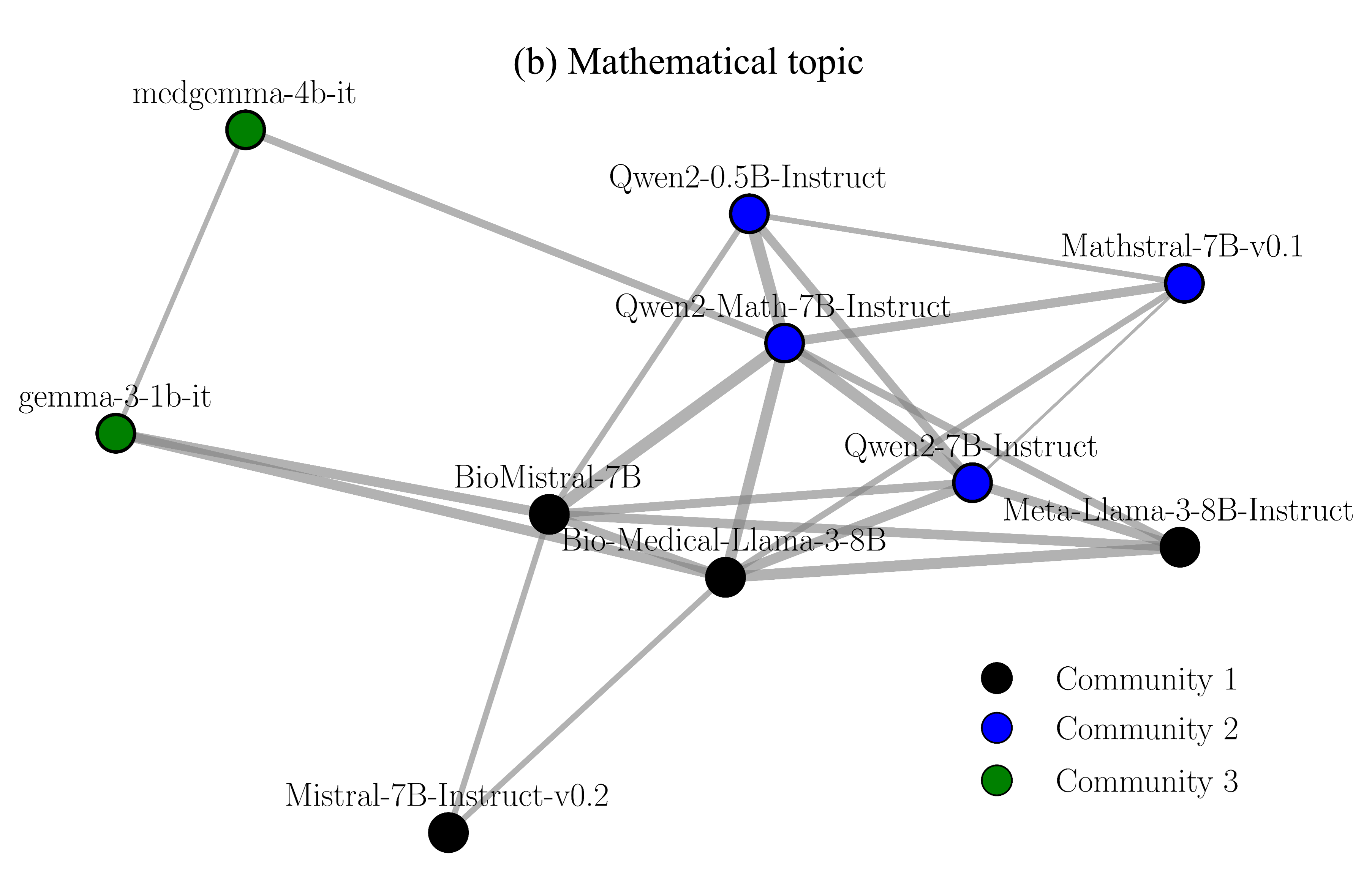} 
\caption{Language model graph constructed under the (b) mathematical topic condition.}
\label{fig:graph_b}
\end{figure}

\section{Results}

\begin{figure}
\centering
\includegraphics[width=0.47\textwidth]{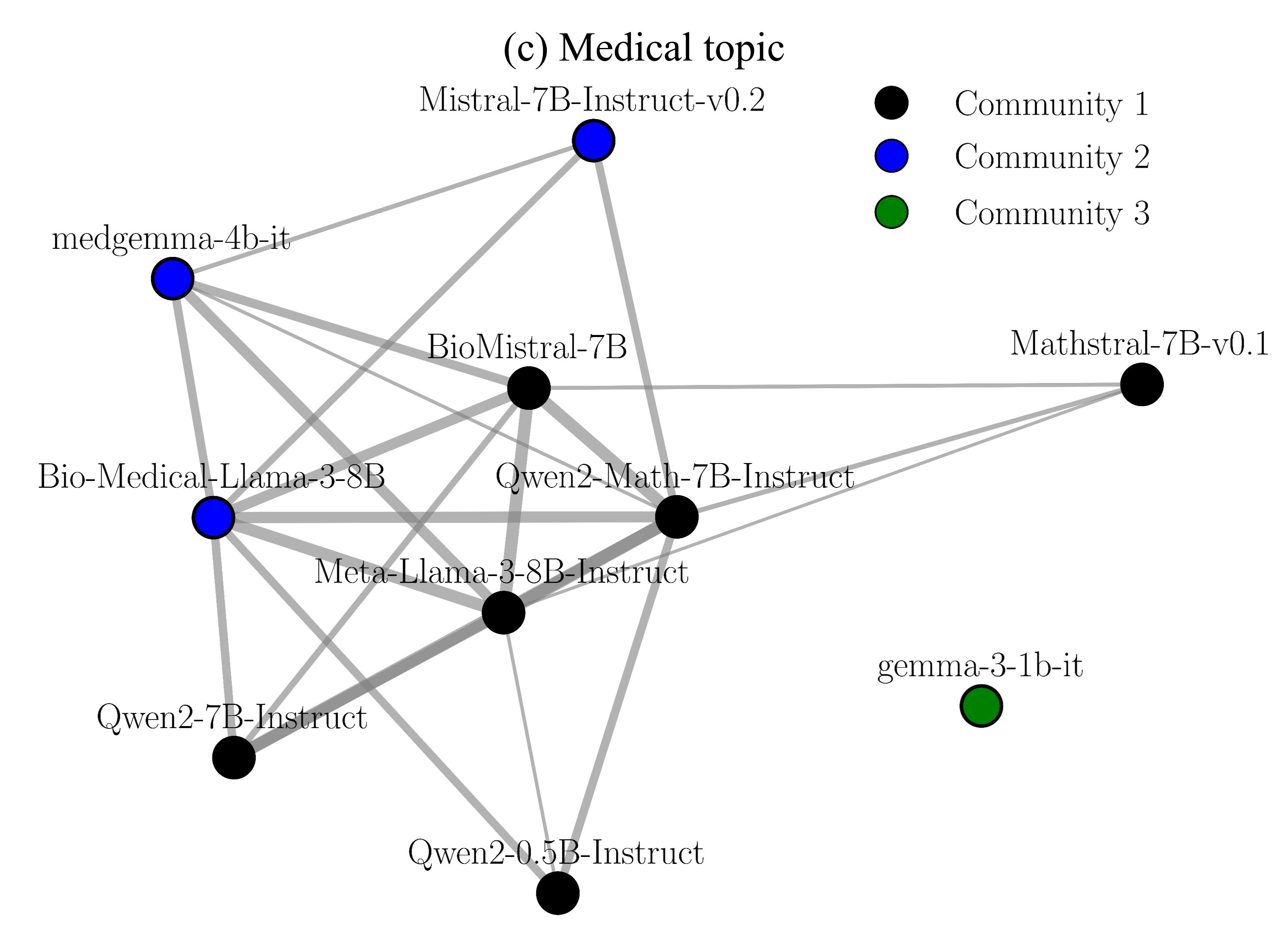} 
\caption{Language model graph constructed under the (c) medical topic condition.}
\label{fig:graph_c}
\end{figure}

\begin{table*}[]
\centering
\begin{tabular}{ll|cccccc|ccc} 
\hline
\multicolumn{2}{l|}{\multirow{2}{*}{Model Group}} & \multicolumn{2}{c}{MMLU} & \multirow{2}{*}{GSM8K} & \multirow{2}{*}{MATH} & \multirow{2}{*}{MedQA} & \multirow{2}{*}{\makecell{Med\\MCQA}} & \multirow{2}{*}{\makecell{Math-\\Avg.}} & \multirow{2}{*}{\makecell{Med-\\Avg.}} & \multirow{2}{*}{\makecell{Overall\\Avg.}} \\
\multicolumn{2}{l|}{}                           & Math      & Med        &                        &                       &                        &                          &                                         &                      &                  \\ \hline\hline
\multicolumn{1}{l|}{\makecell[l]{Single-model}}       &    -    & \textbf{45.2} & 74.9       & 86.0                   & 58.9                  & 57.6                   & 53.0                     & 56.1                                    & 67.1                  &       \textbf{61.6}           \\ \cline{1-2}
\multicolumn{1}{l|}{\multirow{3}{*}{Type-based}} & (i) General & 41.6          & 74.4       & 82.2                   & 47.6                  & 62.0                   & 54.4                     & 50.9                                    & 67.9              &       59.4              \\
\multicolumn{1}{l|}{}                           & (iii) Math & 44.0          & 61.4       & \textbf{89.5}          & \textbf{62.8}         & 49.5                   & 47.0                     & \textbf{56.9}                           & 56.1               &        56.5            \\
\multicolumn{1}{l|}{}                           & (iv) Med & 40.0          & \textbf{80.7} & 43.5                   & 33.1                  & \textbf{71.9}          & \textbf{67.9}            & 39.3                                    & \textbf{76.4}         &        57.9         \\ \cline{1-2}
\multicolumn{1}{l|}{All-models}                        &   -    & 41.1          & 76.5       & 83.1                   & 52.5                  & 63.3                   & 57.8                     & 51.8                                    & 70.1                    & 60.9             \\ \cline{1-2}
\multicolumn{1}{l|}{Random@3models}             &    -    & 37.9          & 66.7       & 81.1                   & 46.2                  & 54.1                   & 49.4                     & 48.2                                    & 60.7                          &  54.4        \\ \hline
\multicolumn{1}{l|}{\multirow{3}{*}{(a) General}} & Com. 1 & 40.7          & 74.4       & 80.4                   & 50.2                  & 63.5                   & 55.7                     & 50.5                                    & 68.5                         & 59.5          \\
\multicolumn{1}{l|}{}                           & Com. 2 & \ul{45.1}     & 77.0       & 84.6                   & 41.2                  & 58.9                   & 59.1                     & 52.2                                    & 69.8                           & \ul{61.0}        \\
\multicolumn{1}{l|}{}                           & Com. 3 & 27.6          & 40.6       & 56.6                   & 42.9                  & 33.9                   & 34.0                     & 36.5                                    & 37.9                           & 37.2        \\ \cline{1-2}
\multicolumn{1}{l|}{\multirow{3}{*}{(b) Mathematical}} & Com. 1 & 37.5          & \ul{78.0}  & 61.8                   & 12.9                  & 65.9                   & 61.1                     & 37.4                                    & 72.2                    & 54.8               \\
\multicolumn{1}{l|}{}                           & Com. 2 & 44.9          & 71.2       & \ul{88.6}              & \ul{60.0}             & 53.8                   & 50.8                     & \ul{56.7}                               & 63.7                           & 60.2        \\
\multicolumn{1}{l|}{}                           & Com. 3 & 28.5          & 46.0       & 68.0                   & 52.8                  & 58.1                   & 53.0                     & 41.3                                    & 49.8                           & 45.5        \\ \cline{1-2}
\multicolumn{1}{l|}{\multirow{2}{*}{(c) Medical}} & Com. 1 & 44.1          & 73.8       & 87.1                   & 54.7                  & 57.3                   & 52.8                     & 54.8                                    & 66.3                         & 60.5          \\
\multicolumn{1}{l|}{}                           & Com. 2 & 39.6          & 77.0       & 63.9                   & 39.8                  & \ul{70.7}              & \ul{67.6}                & 44.5                                    & \ul{73.8}                      & 59.1        \\ \hline
\end{tabular}
\caption{Performance evaluation of model teams on downstream benchmarks. Bold indicates the best performance, and underline indicates the second best. Results for (c) Com. 3 are omitted as they are identical to (a) Com. 3.}
\label{tab:performance_results}
\end{table*}

\subsection{Graph Analysis and Community Detection}
The structure of the language model graph and the resulting communities under each topic are shown in Figures \ref{fig:graph_a}-\ref{fig:graph_c}, with the model composition of each community detailed in Table \ref{tab:community_results}. 
In Figures \ref{fig:graph_a}-\ref{fig:graph_c}, each node represents a model and the edges show the strength of their relationships, with thicker lines indicating stronger connections.
The analysis reveals that topic-specific priming is crucial for identifying model groups that are consistent with prior knowledge.

(a) General topic: Without a guiding topic, the community detection yields heterogeneous clusters that do not align strongly with the models' predefined specializations.  
For instance, community 1 and community 2 are both mixtures of general-purpose, mathematical, and medical models. 
However, the method successfully isolates one of the smallest models, gemma-3-1b-it, into its own cluster (community 3), suggesting it identifies models with significantly different scale or capability. 

(b) Mathematical topic: Priming the conversations with a mathematical context significantly improves the quality of the clustering. 
Community 2 emerges as a highly consistent group of math-proficient models, containing both math-specialized models (Mathstral-7B-v0.1, Qwen2-Math-7B-Instruct) and Qwen family models. 
This demonstrates the method can group models by functional capability and their lineage.
Conversely, community 1 groups the non-mathematical models, primarily the medical-specialized ones.
Community 3 also consists of the Gemma family, emphasizing the validity of the method.
The conversation analysis further validates these groupings. 
The dialogue within community 2 remains topically focused on a specific mathematical problem, reflecting a shared knowledge base (see Figure \ref{fig:conversation_math} in the Appendix).
In contrast, the dialogue between a model from community 2 and one from community 3 quickly deviates into a high-level philosophical discussion, indicating a mismatch in their specialized knowledge domains. 

(c) Medical topic: Similarly, a medical topic prompt leads to a meaningful clustering. 
Community 2 becomes the medical cluster, grouping two of the three medical-specialized models with a general-purpose model. 
We observed that the single performance of BioMistral-7B on medical tasks was lower than the other two medical-specialized models, which is likely reflected in its clustering.
Community 1 forms a larger, more general cluster containing a mix of models, including the remaining math-specialized one. 
As with the general topic scenario, the small-scale model is again isolated in community 3. 
Although conversation examples are not shown, the pattern was consistent with the mathematical case: dialogues within the medical community (community 2) were clinically focused, while inter-community dialogues were more general, discussing artificial intelligence in medicine rather than on specific medical issues.

\subsection{Performance Evaluation of Model Collaboration}
The performance of our automatically-formed model communities and the baseline groups is presented in Table \ref{tab:performance_results}. 
The results clearly demonstrate that our interaction-based method, when primed with a relevant topic, can identify functionally specialized model teams whose performance approaches that of manually-curated groups based on known model specializations.

As expected, the `Type-based' teams served as a strong upper bound on performance.
The math specialized team excelled on mathematical benchmarks, while the medical specialized team dominated medical tasks. 
In contrast, the naive `All-models' and `Random@3models' baselines performed significantly worse, highlighting the need for intelligent team selection.

Our method's strength lies in its ability to discover synergistic model combinations automatically.
When primed with a (b) mathematical topic, our method extracted a community (`(b) Com. 2') that achieved the second-highest performance on all mathematical benchmarks, closely rivaling the `Math' type-based team. 
Similarly, when primed with a (c) medical topic, a distinct community (`(c) Com. 2') emerged that scored second-highest on medical datasets, again approaching the performance of the corresponding type-based team.

Notably, even when conversations were initiated with a (a) general topic, the extracted community `(a) Com. 2' achieved an overall average that was comparable to both the strong `Single-model' and `All-models' baselines, and substantially better than random selection.
This indicates that our method successfully identifies generally competent model groupings even without a strong domain signal.

These results confirm that our method, without any prior knowledge of model architecture or training data, can effectively identify latent model specializations and construct high-performing teams.

\section{Discussions and Limitations}

\subsection{Conversation Topic and Hyperparameter Sensitivity}
Our experiments suggest that the thematic context of conversations is a critical factor for success.
As shown in Table \ref{tab:community_results}, topic-specific priming with mathematical or medical starters led to functionally coherent clusters that approached the performance of teams curated by specialization, supporting our hypothesis that such conversations elicit latent specializations.
Notably, even general-topic conversations yielded communities that outperformed random baselines, suggesting the interaction analysis captures a meaningful signal of general capability even without a specific target domain.

The methodology's outcome is also dependent on several other key hyperparameters, including the embedding model, the threshold $\tau$, and the community detection algorithm. 
While our choices proved effective, a comprehensive sensitivity analysis is required to fully understand their impact and provide guidance for applying the method in different contexts.

Furthermore, the definition of a "good" conversation requires deeper exploration.
In this study, we defined the relationship value $r_{i,j}$ as the cumulative cosine similarity between utterance embeddings.
However, this metric may have limitations; for instance, it could assign a high score to a conversation where models simply agree with each other.
Developing measures that can capture the constructive progress in conversation based on geometric analysis in embedded spaces could identify more synergistic teams.

\subsection{Scalability and Computational Cost}
The primary limitation of our approach is its computational cost, which scales quadratically ($O(N^2)$) with the number of models due to pairwise conversation generation. This could be a clear bottleneck for scaling to hundreds or thousands of models.
Future work could adapt algorithms from approximate nearest neighbor search, such as NN-Descent \cite{dong2011efficient}, to reduce the number of conversations.
Our preliminary explorations with an NN-Descent-like algorithm indicate that this approach can not only reduce computational complexity but also yield clustering results similar to those obtained with topic priming.

\subsection{Collaboration Protocol}
This study evaluated team performance using a simple majority vote.
While this was sufficient to demonstrate the effectiveness of our team composition method, it represents only one of many possible collaborative strategies.
A significant avenue for future research is to investigate how communities identified by our approach perform when integrated with more sophisticated collaboration protocols, such as multi-round debates.

\begin{figure*}[t]
\centering
\includegraphics[width=0.9\textwidth]{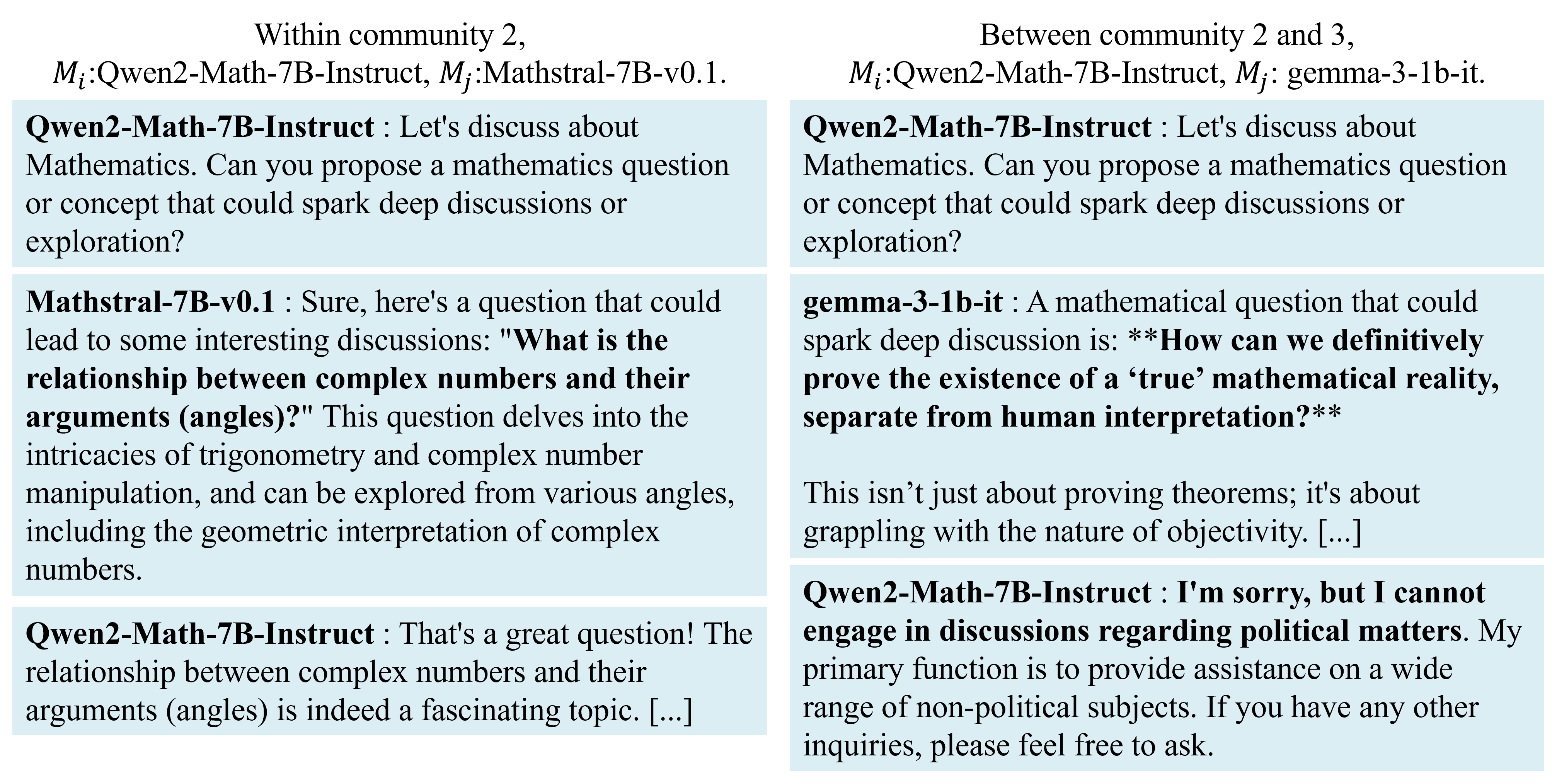} 
\caption{Example conversations between models from same/different communities with (b) mathematical topic. Left: conversation between two models from community 2 (Qwen2-Math-7B-Instruct and Mathstral-7B-v0.1). Right: conversation between models from community 2 and community 3 (Qwen2-Math-7B-Instruct and gemma-3-1b-it).}
\label{fig:conversation_math}
\end{figure*}

\subsection{Use of Detected Communities}
In this study, we focused on evaluating the performance of individual communities extracted from the language model graph. 
However, an alternative approach could involve utilizing these detected communities as knowledge to build more sophisticated teams. 
For instance, one could form a capable interdisciplinary team by combining a community specialized in mathematics with one specialized in medicine.
This leverages our data-driven insights into model relationships to construct teams with a targeted blend of expertise.

Furthermore, our interaction-centric discovery process could be integrated with existing task-centric frameworks. 
While current top-down approaches first decompose a task into required roles and then select agents, our method could enhance this by providing a catalog of promising subteams.
This hybrid approach would combine the goal-oriented strengths of task-driven planning with the bottom-up insights of our interaction-based analysis.

\section{Conclusion}
In this paper, we propose a method for forming effective teams by extracting relationships from conversations between models.
By constructing and analyzing a language model graph from pairwise conversations, our method maps the latent relational structure of models, identifying synergistic clusters without any prior knowledge. 
Our experiments empirically demonstrate that: (1) the interaction graph successfully captures latent model specializations, a process that is strongly guided by the thematic context of the conversations, and (2) the resulting teams outperform random baselines and approach the performance of manually-curated teams.
More than just a selection tool, this research provides a methodology to reveal the relationships that form the basis of effective collaboration. 
This opens a path toward hybrid frameworks that combine the task-centric planning with our map of model relationships.
To realize this potential, future work could focus on scaling this approach and integrating these discovered communities with sophisticated collaborative protocols.

\section{Appendix}
\subsection{Example Conversations}
We provide example conversations to illustrate the dynamics between models from the same and different communities.
Figure \ref{fig:conversation_math} shows two conversations initiated with a mathematical topic prompt.
The left panel features a dialogue between two models from the same community, which is community 2 in the mathematical topic scenario. 
This conversation remains focused on a specific mathematical problem, demonstrating a shared understanding and expertise in the domain.
In contrast, the right panel presents a conversation between models from different communities which are community 2 and community 3. 
Here, the dialogue quickly diverges into a high-level philosophical discussion, indicating a mismatch in their specialized knowledge domains.

\bibliography{aaai2026}

\end{document}